\documentclass[letterpaper, 10 pt, conference]{ieeeconf}  %
\IEEEoverridecommandlockouts                              %
\usepackage{graphicx}
\usepackage{amsmath}
\usepackage{amssymb}
\usepackage{booktabs}
\usepackage{dsfont}
\usepackage{times}
\usepackage{epsfig}
\usepackage{graphicx}
\usepackage{ctable}
\usepackage{enumerate}
\usepackage{subcaption}
\usepackage{microtype}
\usepackage{color}
\definecolor{darkred}{rgb}{0.5, 0.0, 0.0}
\definecolor{darkblue}{rgb}{0.0, 0.0, 1.0}
\definecolor{darkgreen}{rgb}{0.0, 0.5, 0.0}
\usepackage[pagebackref,breaklinks,colorlinks]{hyperref}
\usepackage[capitalize]{cleveref}
\crefname{section}{Sec.}{Secs.}
\Crefname{section}{Section}{Sections}
\Crefname{table}{Table}{Tables}
\crefname{table}{Tab.}{Tabs.}

\title{\LARGE \bf
Instance Segmentation with Cross-Modal Consistency
}

\author{Alex Zihao Zhu$^{1}$, Vincent Casser$^{1}$, Reza Mahjourian$^{1}$, Henrik Kretzschmar$^{1}$ and S\"oren Pirk$^{2}$%
\thanks{$^{1}$Waymo LLC}%
\thanks{$^{2}$Adobe Research (Work done while at Google Research)}%
}

\begin{document}

\thispagestyle{empty}
\pagestyle{empty}
\maketitle

\begin{abstract}
Segmenting object instances is a key task in machine perception, with safety-critical applications in robotics and autonomous driving. We introduce a novel approach to instance segmentation that jointly leverages measurements from multiple sensor modalities, such as cameras and LiDAR.
Our method learns to predict embeddings for each pixel or point that give rise to a dense segmentation of the scene. Specifically, our technique applies contrastive learning to points in the scene both across sensor modalities and the temporal domain.
We demonstrate that this formulation encourages the models to learn embeddings that are invariant to viewpoint variations and consistent across sensor modalities. We further demonstrate that the embeddings are stable over time as objects move around the scene. This not only provides stable instance masks, but can also provide valuable signals to downstream tasks, such as object tracking.
We evaluate our method on the Cityscapes and KITTI-360 datasets. We further conduct a number of ablation studies, demonstrating benefits when applying additional inputs for the contrastive loss.
\end{abstract}

\section{Introduction}

Identifying and segmenting objects is a fundamental and challenging task in computer vision, and critically important for many robotics applications, such as autonomous driving. Given some sensor input such as a camera image or LiDAR scan, the goal of instance segmentation is to associate each pixel or point with a unique instance label for the object to which it corresponds. This allows for fine-grained reasoning about each object beyond sparse representations such as bounding boxes. Only recently, learning-based approaches have shown remarkable success in solving this task on a number of benchmark datasets~\cite{Cordts_2016_CVPR,nuscenes2019,lin2014microsoft}. Reliably identifying object instances in complex real-world data, both across different data modalities and coherent in time, %
remains a challenging and open problem~\cite{Xiong_2019_CVPR,cheng2020panoptic,DBLP:conf/eccv/ChenLCCCZAS20}.

\begin{figure}[t]
  \begin{center}
  \includegraphics[width=\linewidth]{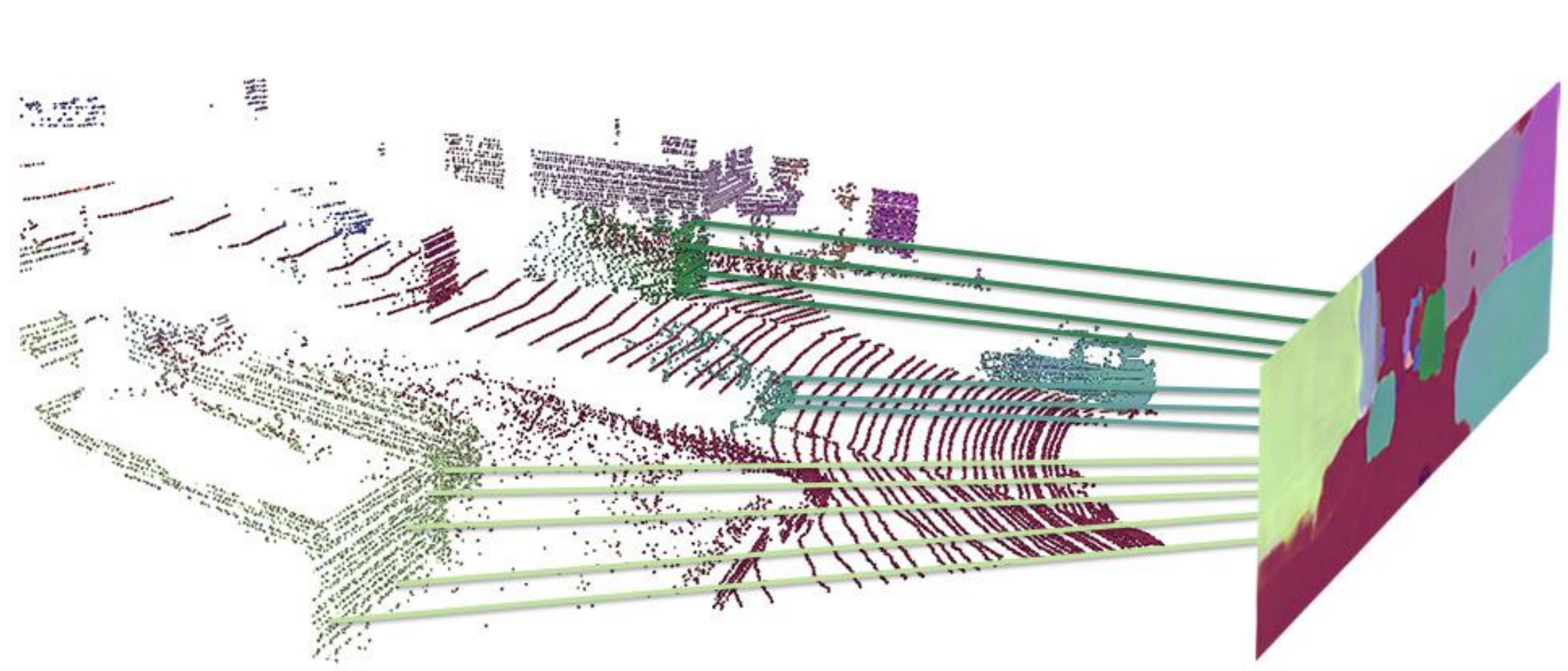} 
  \end{center}
  \vspace{-4mm}
  \caption{Our approach to instance segmentation leverages information from multiple sensor modalities, such as 3D LiDAR point clouds and camera images (left). Our method learns to predict dense embeddings (depicted by the colors, right image) based on a novel cross-modal contrastive loss that uses samples from different sensor modalities and temporally consecutive frames of RGB images and LiDAR points. The resulting embeddings are stable over time and invariant to viewpoint changes.}
  \label{fig:network_overview_multimodal}   
  \vspace{-5mm}
\end{figure}

\begin{figure*}[ht!]
  \begin{center}
  \includegraphics[width=\linewidth]{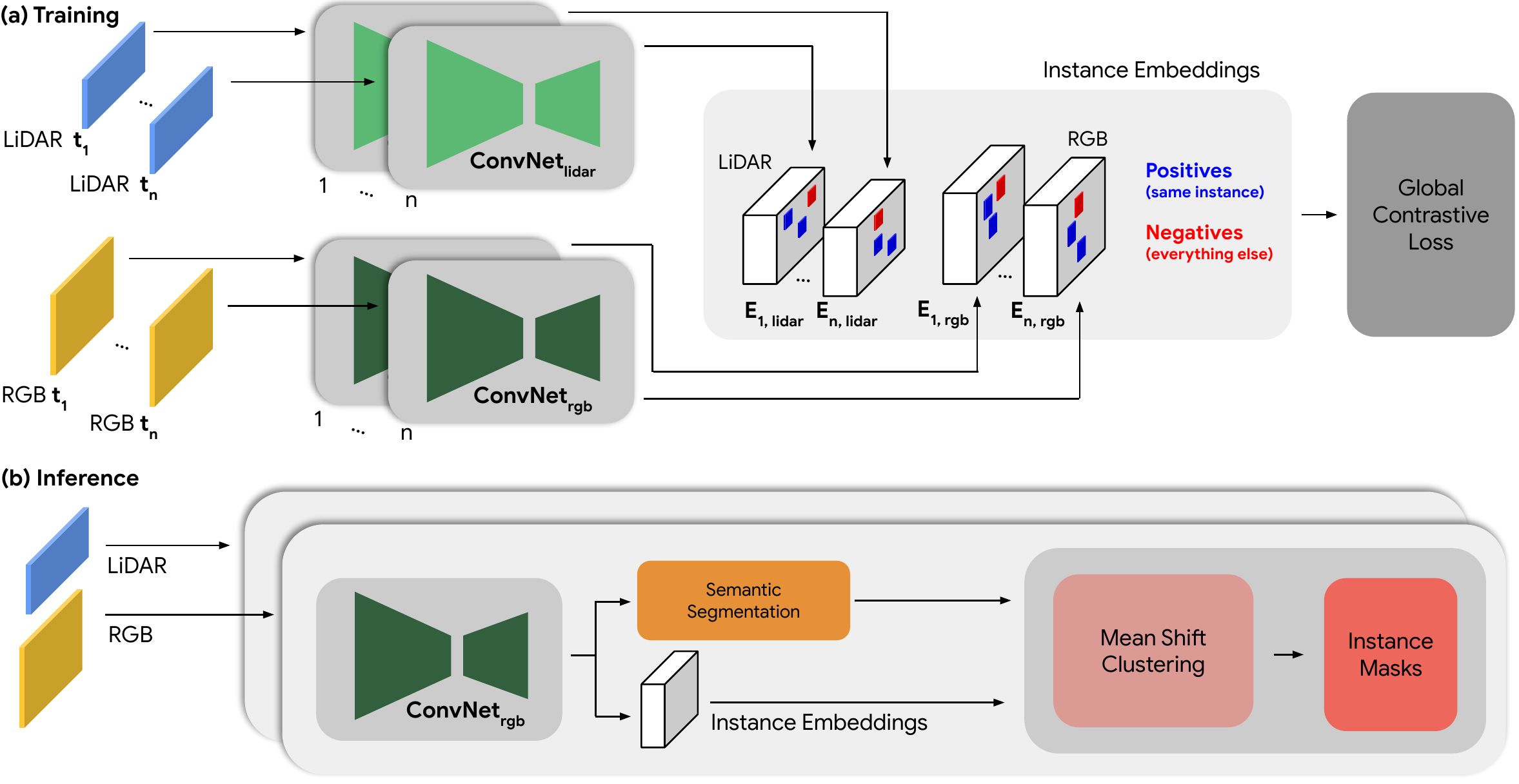} 
  \end{center}
  \vspace{-3mm} %
  \caption{Our goal is to learn dense cross-modal embeddings via contrastive learning to segment individual object instances. Our method leverages shared information between sensors and consecutive frames by applying the contrastive loss jointly over embeddings from all input sources at training time. This global contrastive loss allows the network to share information between different data sources, and improves overall segmentation performance. Once trained, the learned embeddings can be used to obtain object IDs through mean shift clustering.}
  \label{fig:network_overview}   
  \vspace{-5mm}
\end{figure*}

Many safety-critical robotics systems, such as autonomous vehicles, rely on multiple sensors that provide complementary information to perceive the environment. For instance, LiDAR sensors provide highly accurate range readings, but only in clear weather conditions and up to a limited range. On the other hand, cameras may be able to perceive objects at much longer ranges, but without any explicit range readings and only in adequate lighting conditions. As a result, there is an opportunity to improve segmentation performance for each sensor by reasoning jointly across all modalities.

Another avenue is the utilization of temporal data, which can provide motion information, varying viewpoints and context for occlusions. Temporal data comes naturally for most sensors, although sequential labeling for ground truth is more expensive.

In this paper, we aim to leverage these complementary signals and advance instance segmentation by training jointly on different data modalities and over time (Figure~\ref{fig:network_overview_multimodal}). Specifically, we propose a novel cross-modal contrastive loss that enables us to train with sequential samples from camera images and LiDAR point clouds to learn a dense representation of object instances. Our goal is to learn embeddings that coherently represent individual objects in both modalities. By training on different modalities, our method can more meaningfully disentangle individual instances. Moreover, we show that our loss formulation is flexible enough to be applied to sequences of sensor data of both modalities, enabling learning temporally-stable embeddings for object instances. For situations where only sequential data (but no labels) are available, we propose a method for generating pseudo-groundtruth using optical flow, and demonstrate that training on this data brings similar improvements in segmentation.

In summary, our contributions are as follows: 
\begin{itemize}
\itemsep0em 
\item
We introduce a novel cross-modal contrastive loss to obtain consistent instance embeddings for RGB images and point clouds.
\item
We show that our loss can also be used to train on \textit{sequences} of sensor data from both modalities, enabling the learning of temporally stable embeddings.
\item
We propose a method using optical flow to generate pseudo ground-truth for video, allowing the extension of our contrastive loss to partially labeled datasets.
\item
Our extensive experiments on the KITTI-360~\cite{liao2021kitti} and Cityscapes~\cite{Cordts_2016_CVPR} benchmarks suggest that our method significantly improves instance segmentation quality by up to 1.9~mAP when contrasting between temporal RGB data, and by up to 1.7~mAP when contrasting between temporal LiDAR and RGB data, while also improving the temporal stability of the embeddings.
\end{itemize}

\vspace{2mm}
\section{Related Work}

Object detection and segmentation have a long tradition in computer vision. While this spans a wide range of different approaches that we cannot conclusively discuss, we aim to provide an overview of learning-based methods toward instance and semantic segmentation in images and point clouds.

\textbf{Semantic Segmentation.} The goal of semantic segmentation is to associate every pixel in an image with a discrete label that defines its semantic class. 
A major challenge of effectively training networks for semantic segmentation is to define precise labels, and a number of different datasets were introduced for benchmarking~\cite{Cordts_2016_CVPR,Geiger2013IJRR,nuscenes2019}. Many recent methods for semantic segmentation employ deep neural network architectures for feature aggregation based on pyramid pooling~\cite{8100143}, attention modules~\cite{Chen2016AttentionTS}, multi-scale activations~\cite{Hariharan_2015_CVPR}, or end-to-end convolutions~\cite{7298965}. Furthermore, it has been recognized that dilated or atrous convolutions provide a powerful way of capturing more global features through wider receptive fields~\cite{8099558,journals/corr/ChenPK0Y16}. Other approaches aim to solve semantic segmentation with an emphasis on learning structured representations based on conditional random fields~\cite{Arnab2016HigherOC,journals/corr/ChenPK0Y16} or deep parsing networks~\cite{10.1109/ICCV.2015.162}.

\textbf{Instance Segmentation} expands on the concept of semantic segmentation. Here the goal is to not only obtain a class label for every pixel, but to predict labels for individual object instances. Instance segmentation approaches can be separated into two different categories. In the first category, most methods rely on a two stage (or top-down) process~\cite{10.1007/978-3-319-46466-4_32,7299025,he2017mask}. A region proposal step \cite{6909475} identifies object instances in a first stage and each region is then segmented into foreground and background during a second stage. The performance of top-down approaches is determined by obtaining region proposals and by the number of objects present in an image. On the contrary, bottom-up approaches  aim to learn per-pixel embeddings that represent individual object instances based on recurrent instance grouping~\cite{kong2018grouppixels}, watershed transformations~\cite{8099788}, subnetwork instantiations~\cite{arnab2017}, associative embeddings~\cite{10.5555/3294771.3294988}, or based on more refined loss functions such as discriminative loss~\cite{Brabandere2017SemanticIS,hu2019learning}, or metric learning~\cite{Fathi2017SemanticIS}. 
Bottom-up approaches are commonly implemented with more lightweight architectures, which offers more efficient training and easier integration into existing multi-task setups. However, learning meaningful embeddings from complex data to represent individual objects remains a challenging problem. Finally, a number of hybrid approaches exist that aim to learn embeddings of larger regions in images~\cite{chen2020blendmask,chen2019}.

\textbf{3D Object Detection and Segmentation:} Detecting and segmenting objects in point clouds has received a considerable amount of attention. PointNets~\cite{qi2016pointnet} directly consume raw point clouds to generate latent representations that can be used for classification or semantic segmentation tasks. By leveraging various geometric relationships and representations a number of methods have since then improved upon the performance to solve for tasks, such as object detection~\cite{Zhou2018VoxelNetEL}, semantic segmentation~\cite{8578577,NIPS2017_d8bf84be}, or instance segmentation~\cite{8953321,Yi2019GSPNGS}. A number of approaches rely on Bird's Eye View (BEV) representations to generate bounding boxes~\cite{Yang2018PIXORR3,chen2017-multi-view} that -- along with pillar-based representations~\cite{Lang2019PointPillarsFE,Wang2020PillarbasedOD} -- are popular in autonomous driving. More recent methods for 3D instance segmentation rely on more principled approaches. To this end 3D-MPA ~\cite{Engelmann20203DMPAMA} uses an aggregation-based approach where each point votes for its object center,  HAIS~\cite{chen2021hierarchical} exploits the spatial relation of points and point sets, and SSTNet~\cite{liang2021instance} uses a hierarchical representation which is constructed based on learned semantic features.

\textbf{Multimodal Approaches.} 
There also exists a limited number of multimodal approaches proposed in the robotic-vision domain, including~\cite{xiang2020learning,meyer2020improving}. Most of these works tackle indoor environments using closely paired RGB-D data.

Similar to these methods, our goal is to leverage multimodal sensor data. Specifically, we use a single-stage fully-convolutional network to learn per-point and per-pixel embeddings for instance segmentation. We train these embeddings with a metric learning loss that contrasts embeddings of the same object instance with those of different objects. The key idea to our approach is that we obtain the positive and negatives samples spatially, within the same frame, across different modalities (images, points), and even across time, in the adjacent frames of a sequence. We accomplish this by proposing a novel cross-modal consistency loss that allows us to sample embedding vectors from different sources. Training a single-stage network with this loss enables us to compute dense embeddings and, consequently, more robust instance masks for each frame, while also obtaining temporally stable results for sequences.

\section{Method}

In this section, we introduce our model architecture, the cross-modal contrastive loss function, and implementation details. Our full pipeline is shown in Figure~\ref{fig:network_overview}. 

\subsection{Cross-Modal Contrastive Loss}

Our method consists of three main components: (1) a pair of instance segmentation networks that generate dense embeddings for RGB images and LiDAR range images. (2) a sampling based contrastive learning loss which constrains embeddings from the same instances to be similar, while forcing embeddings from different instances to be different. (3) a consistency module based on paired instance labels that introduces additional training data and correspondences between sensors and over temporally neighboring frames.

\textbf{Contrastive Loss:} The goal of the instance segmentation network is to segment object instances by learning dense embedding vectors. The network takes as input either RGB images or LiDAR range images~\cite{bewley2020range}, and outputs embeddings with $c=32$ channels. To train the network, we use a metric learning loss for contrasting embeddings of the same instance with those of different instances. Embeddings of the same instance are pushed closer together during training, while those of different instances are pushed away in the embedding space.

\begin{figure}[t]
  \begin{center}
  \includegraphics[width=\linewidth]{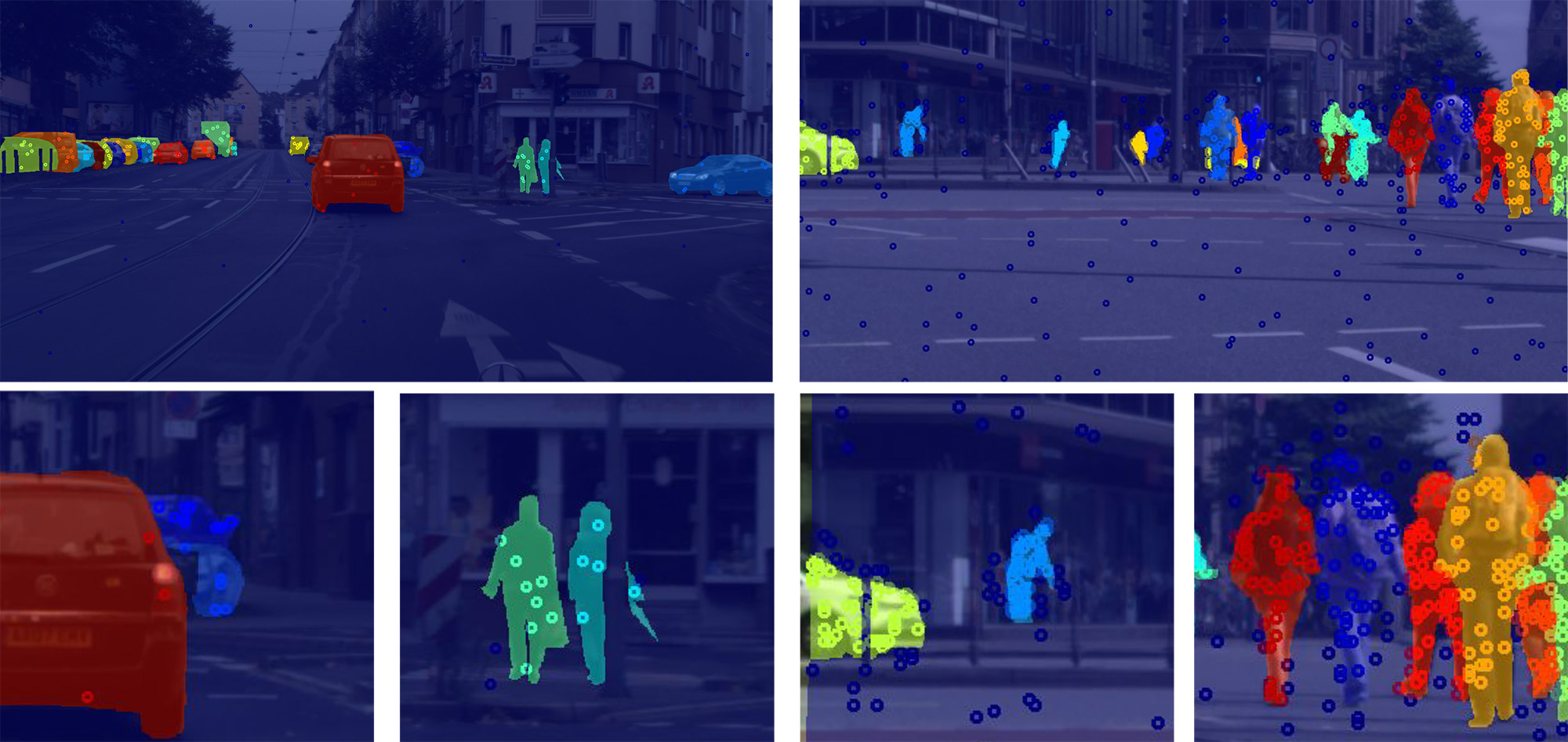} 
  \end{center}
  \vspace{-3mm}  %
  \caption{Example of our sampling strategy where points are evenly distributed amongst instances. The bottom row shows close-ups of the used sample locations.}  
  \label{fig:sampling_strategy}   
\end{figure}

\begin{figure}[t]
  \begin{center}
  \includegraphics[width=\linewidth]{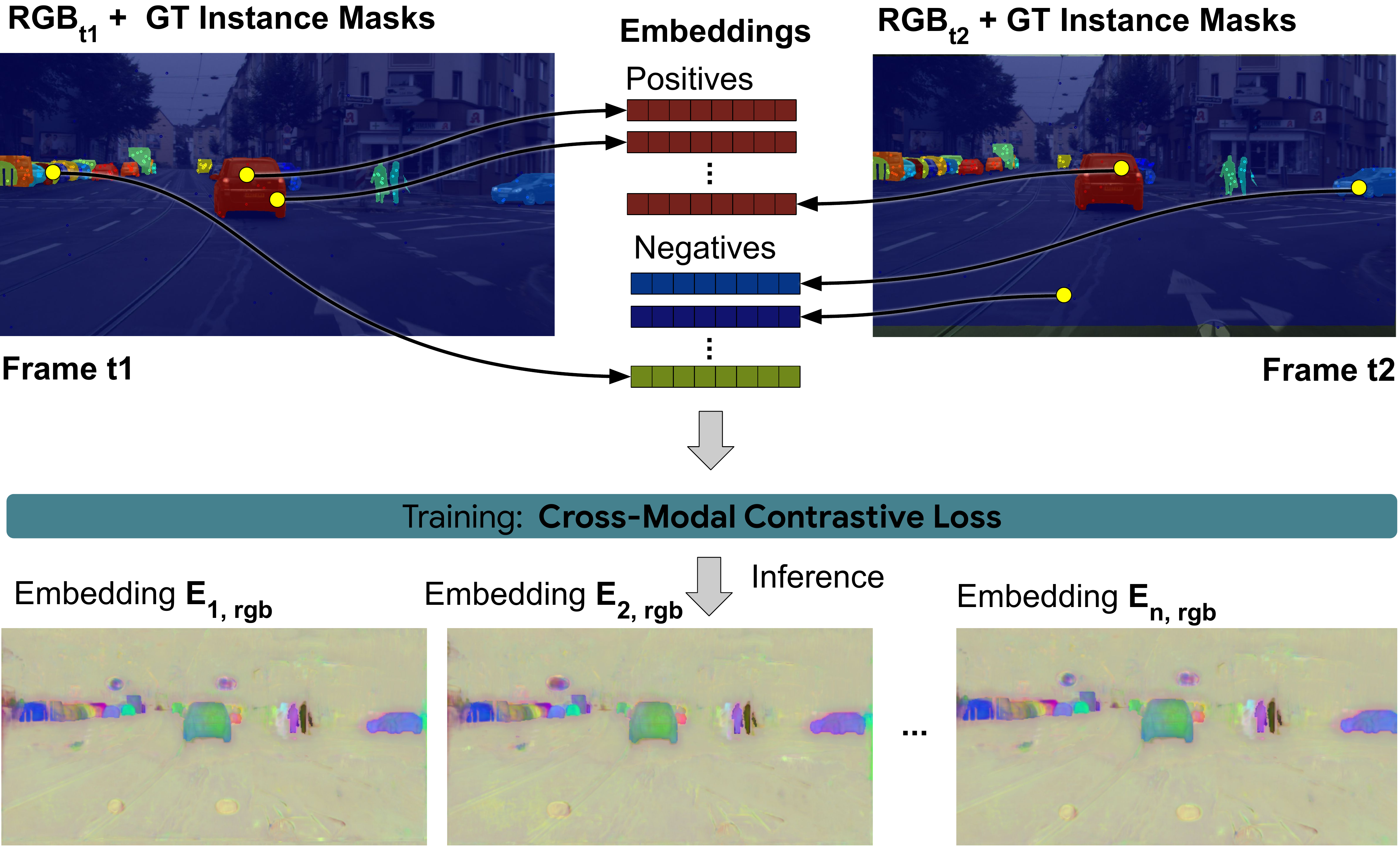} 
  \end{center}
  \vspace{-3mm}  %
  \caption{Given a set of predicted embeddings with paired instance labels, we sample positive embedding vectors within the instance mask of an object in the current frame as well in all other frames (cross sensor and temporal). Samples from all other objects in all frames are used as negatives. 
  We then train with our cross-modal contrastive loss to learn dense embeddings of object instances. The images at the bottom show the projected embeddings during training.}
  \label{fig:embedding_vis}   
  \vspace{-3mm}   
\end{figure}

Our instance segmentation loss consists of the normalized temperature-scaled cross entropy (NT-XENT) loss defined in~\cite{chen2020simple}. We define the per-pixel embeddings as $E_{h\times w\times c}$, where $h$ and $w$ denote the spatial dimensions of the output, and $c$, the channels of the embedding at each pixel. For each frame, we randomly sample $K_1=8192$ embeddings, distributed evenly across all instances (visualized for RGB frames in Figure~\ref{fig:sampling_strategy}). Given a set of sampled embeddings for an object instance, we generate all possible positive pairs. The samples of all other instances are used as negatives for each pair. For a set of samples $U$ and a positive pair, $e_i, e_j$, our loss can then be defined as:

\begin{align}
l_{i, j}=&-\log\frac{\exp(\text{sim}(e_i,e_j)/\tau)}{\sum_{k\in U}\mathds{1}_{[id(i)\neq id(k))]}\exp(\text{sim}(e_i, e_k)/\tau)}\\
\text{with }&\text{sim}(x, y)=\frac{x^Ty}{\|x\|_2\|y\|_2},
\end{align}

where $\tau$ denotes a temperature hyperparameter and $id(i)$ is the mask ID of point $i$. The similarity function $\text{sim}$ is used to compute the cosine similarity between two normalized embedding vectors $x$ and $y$. The total loss over the instance embeddings is then defined as:
\begin{align}
    l_{\text{c}}=&\frac{1}{\|U\|}\sum_{i, j\in U}\mathds{1}_{[id(i)=id(j)]}l_{i, j}.
\end{align}
Furthermore, to obtain more stable results, we also apply a regularization loss over the norm of the embeddings:
\begin{align}
l_{\text{r}}=&\frac{1}{\|h \times w\|}\sum_{x,y} \|E(x,y)\|_2.
\end{align}
Our final loss, then, is:
\begin{align}
l =& l_c + \lambda l_r, 
\label{eq:final_loss}
\end{align}
where $\lambda=0.01$ is a weighting factor.
Together, these losses allow us to learn meaningful embedding vectors as representation for object instances. 

\textbf{Consistency Over Modalities and Time:}
Given a dataset with instance labels which are consistent between sensors and across time, such that an object has the same ID across all inputs, one can apply the above training loss~\eqref{eq:final_loss} between all inputs by randomly sampling points from each set of embeddings, and using the consistent instance labels to determine positive and negative pairs. This allows the network to learn embeddings which are consistent not only within a single frame, but between all modalities and time steps seen at training.

However, it is often the case that we do not have consistent instance labels. In these cases, we propose a method for generating pseudo-groundtruth using optical flow. Given a dataset with individually labeled frames amongst unlabeled temporal sequences, we use a pre-trained optical flow network to warp the groundtruth of the labeled frame to its temporal neighbors.

We use the predicted flow to warp the groundtruth instance and semantic labels of the current frame to the next frame and apply nearest neighbor resampling to avoid interpolating between integer label values. Following the work of Wang et al.~\cite{wang2018occlusion}, we compute occlusions according to the `range map', defined by the number of points in the original image map to the other image. The instance and semantic labels of occluded pixels, as well as pixels that have left the image, are set to invalid, and are not sampled for the contrastive loss. The embeddings and warped labels for the next frame are then combined with those of the current frame when performing the sampling for the contrastive loss. Figure~\ref{fig:embedding_vis} illustrates the result of the warped labels as well as the sampling of embeddings from the two frames.

\begin{figure}[t]
\begin{center}
\includegraphics[width=\linewidth]{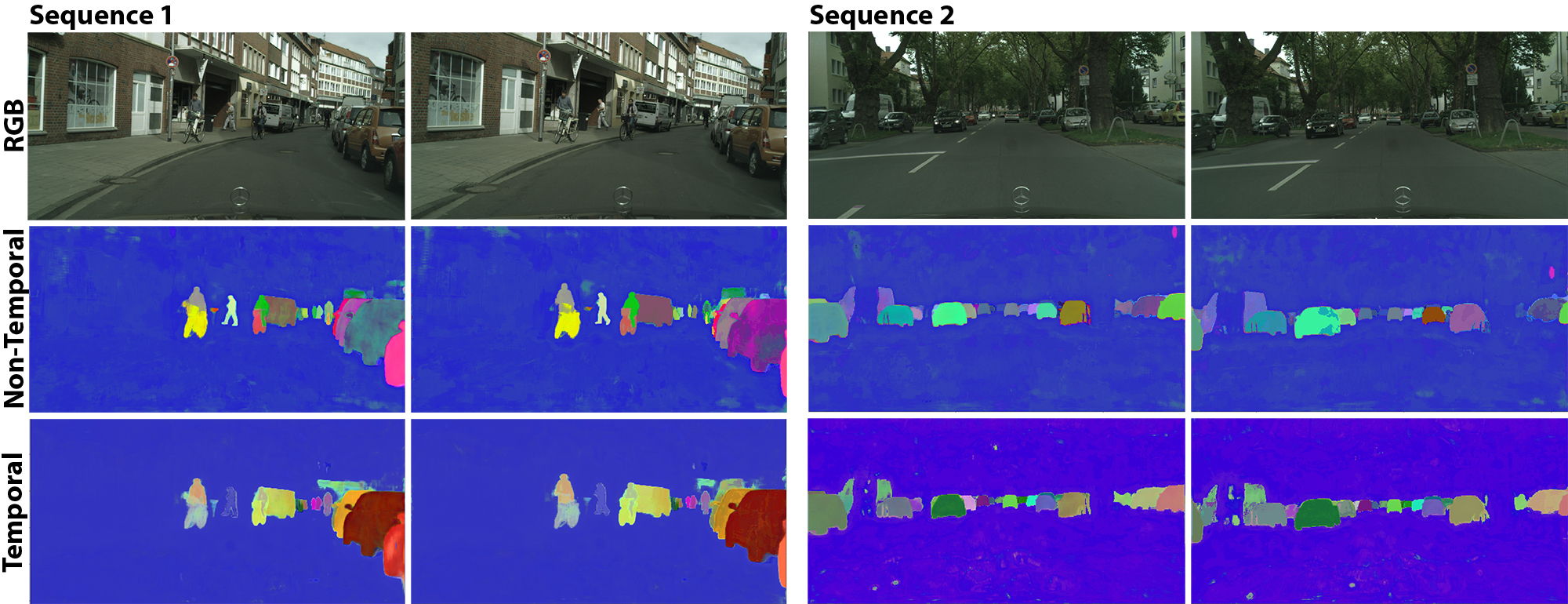} 
\end{center}
\vspace{-3mm}  %
\caption{Temporally-coherent embeddings: as our model is trained with a temporal consistency loss based on optical flow, we are able to obtain temporally-coherent and more stable instance embeddings. Here we show the RGB sequences of two different scenes along with the multi-dimensional per-pixel embedding vectors projected into a color space.}  
\label{fig:temporal_coherence}   
\vspace{-7mm}
\end{figure}

This pipeline has similarities to semi-supervised methods such as in the work by Chen et al.~\cite{DBLP:conf/eccv/ChenLCCCZAS20}, which generate pseudo-groundtruth by running inference with a pre-trained model, and subsequently re-training. However, in addition to providing ground truth to temporally neighboring frames with optical flow, our method also ensures consistency  between frames and thus over time. Sample outputs for our method over time can be found in Figure~\ref{fig:temporal_coherence}.

\textbf{Semantic Segmentation:} We predict a semantic mask for each frame and filter out all `stuff' classes to a single background class. During clustering, we assign semantic classes to each clustered instance with a `majority vote' between semantic predictions within the instance. We also generate the confidence for each mask using the average semantic score. Pixels not belonging to the selected semantic class are dropped from the instance to preserve semantic boundaries. The one-hot semantic labels are concatenated to the predicted instance embeddings, so that the network only has to predict unique instance embeddings within each class.\\

\textbf{Instance Mask Generation:} %

At inference time, we cluster the resulting per-pixel embeddings by applying a variant of the mean shift algorithm proposed by Brabandere et al.~\cite{Brabandere2017SemanticIS}. In particular, we randomly sample a point in the embedding space, and find all inlier points with cosine distance (scaled to be within $[0, 1]$) less than a threshold, $m=0.1$, from the sampled point. We then iterate by shifting the sampled point to the mean of the set of inliers, and repeat until convergence or some maximum number of iterations is reached. We repeat this process until all pixels have been clustered. With this method, we found that erroneous masks were typically generated on the transitions between masks, generating thin artifacts along the boundaries. To compensate, we filter out masks with an area to perimeter ratio less than a threshold, $r=4$. As this method computes distances across the entire image, no assumptions about the connectivity of each mask are made, and so arbitrarily distributed instances can be detected.

More complex methods are available for generating instance masks from embeddings, such as the graph cut algorithm proposed in SSAP~\cite{journals/corr/abs-1909-01616} or the transformer head proposed in MaX-DeepLab~\cite{wang2021max}, which would likely achieve higher performance. However, these would incur additional latency penalties, and their computational power may hide improvements to the underlying embeddings.

\begin{table*} [t]
\centering
\begin{tabular}{l|c|c|c|c|c|c|c|c}
\specialrule{.1em}{.05em}{.05em} 
Method & All & Bicycle & Building & Car & Motorcycle & Person & Rider & Truck \\
\specialrule{.1em}{.05em}{.05em}
Single camera & 11.2 & 0.4 & 22.9 & 42.5 & 1.3 & 5.8 & 0.8 & 4.8\\
Temporal camera & 11.9 & 1.4 & \textbf{23.6} & \textbf{43.1} & 2.0 & 6.5 & 0.8 & 6.1\\
Single camera + LiDAR & 11.9 & 0.4 & 22.5 & 41.8 & 3.1 & \textbf{6.9} & 1.4 & 7.1\\
Temporal camera + LiDAR & 12.2 & \textbf{2.1} & 22.7 & 41.0 & 3.0 & 6.4 & 2.9 & 7.2\\
Temporal camera + Temporal LiDAR & \textbf{12.9} & 2.0 & 22.7 & 40.7 & \textbf{5.4} & 6.3 & \textbf{4.4} & \textbf{8.7} \\
\hline
\end{tabular}
\caption{Ablation study on the effects of additional consistency on instance segmentation performance on a custom KITTI 360 train/val split. Adding additional inputs for the contrastive loss improves overall segmentation performance. Performance is measured in terms of AP ($\%$).}
\label{tab:kitti_ablation}
\end{table*}

\begin{figure*}[t]
  \begin{center}
  \includegraphics[width=\linewidth]{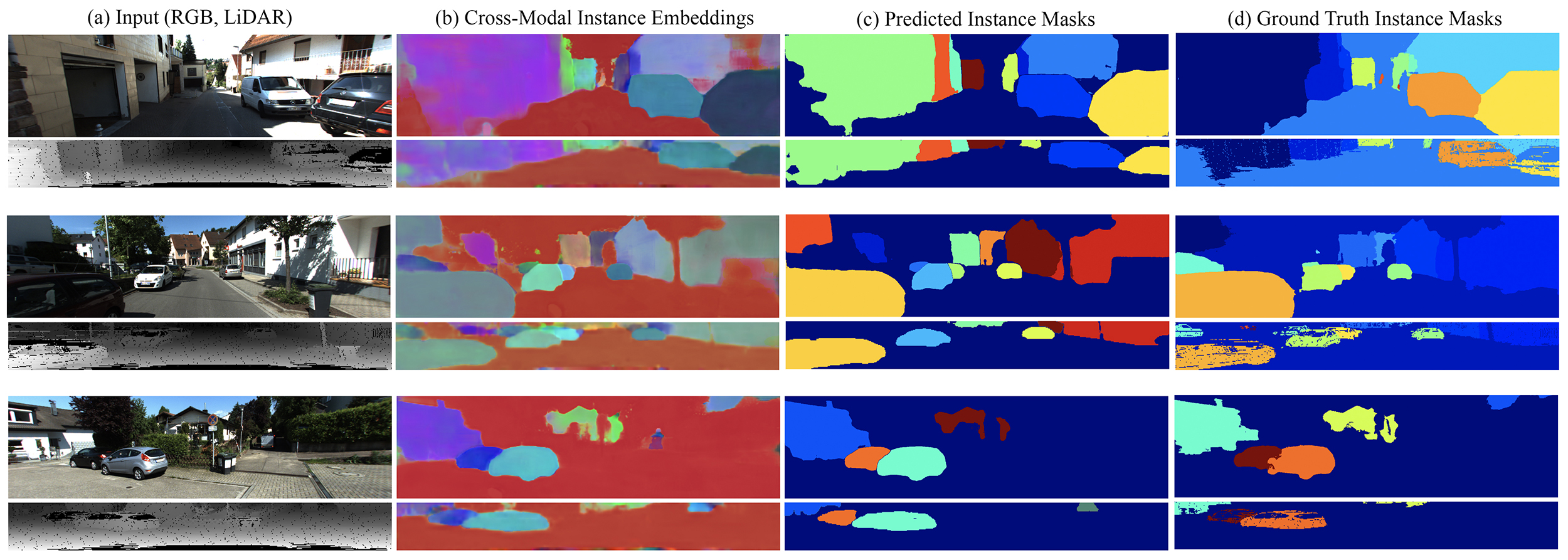} 
  \end{center}
  \vspace{-3mm}  %
  \caption{Cross-modal results on KITTI-360: Our models take as inputs RGB images or LiDAR range images (a), and predicts dense instance embeddings (b) which are consistent between the sensor modalities. At post-processing, we can apply a mean-shift clustering algorithm to generate instance masks for each modality (c), which we compare to the ground truth (d). Our cross-modal contrastive loss contrasts pairs between the sensor modalities, and trains the network to predict the same embedding for each object instance in each sensor, denoted by similarity in color. Each image pair shows both sensor modalities, where the LiDAR data is shown as range images (bottom).}  
  \label{fig:kitti_qualitative}   
\end{figure*}

\subsection{Implementation Details}

In this section, we provide details about the implementation of our network architecture. 

\textbf{Model Architecture:} We use a modified architecture based on Panoptic DeepLab~\cite{cheng2020panoptic}. Our model consists of a Xception-71 backbone, with an additional ASPP and decoder module for instance embedding prediction, with $c=32$ channels, in place of the instance center and offset decoder in ~\cite{cheng2020panoptic}. For our cross-modal experiments, we use two separate networks for LiDAR range image inputs and RGB image inputs.

\textbf{Preprocessing:}
For all datasets, we use full resolution images as input. 
During training, we augment our data by randomly flipping, scaling and cropping or padding the input images. Furthermore, we use random Gaussian blur and jitter brightness, contrast, saturation, and hue. For temporal data, we apply the same augmentation to each pair of images. For LiDAR range images, we only apply flipping, and ensure that the flipping is consistent with the camera images.

\textbf{Training:}
For all experiments, we use a moving average ADAM optimizer with linear warmup for 5 epochs, initial learning rate of 2.5e-4 and exponential decay with decay factor of 1.2 every 10 epochs. For the KITTI-360 dataset, we train for 10,000 steps for the ablation train/val split, and 20,000 steps for the train/test split, with an effective batch size of 128. For the Cityscapes dataset, we train for 60,000 iterations with an effective batch size of 32.

\section{Experiments}

In this section we present and discuss a variety of  experiments, including baseline comparisons, ablation studies, and qualitative results our method is able to generate. For all of our experiments, we do not perform any kind of test time augmentation.

\subsection{Datasets}
\paragraph{KITTI-360}
The KITTI-360 dataset~\cite{liao2021kitti} consists of camera and LiDAR panoptic segmentation and LiDAR bounding box labels for all of the sequences of the KITTI dataset~\cite{Geiger2013IJRR}. For each sequence, the instance labels are consistent both between the camera and LiDAR labels, as well as between different time steps. The dataset contains 83,000 frames and associated LiDAR scans, split into 9 training sequences and 2 test sequences with held out groundtruth. As no validation set is provided, we perform our ablations on a train/val split where sequences 00, 02, 03, 04, 05, 06, 07 are the train set and sequences 09 and 10 are the validation set. The validation set was chosen from the two densest sequences, such that there are almost the same number of objects in the training and validation sets. For the final test set evaluation, we train on the entire training set.

As the dataset is largely transferred from primitive 3D shapes (cuboids, ellipsoids etc.), each label has a corresponding confidence value, which is used to weight the evaluation metrics. In accordance with the experiments in KITTI-360, we only use the top 70\% of labels in each frame, in terms of confidence. To generate temporal frames, we randomly sample an image 2 frames before or after the current frame at each training step. To train a model on LiDAR, we use the range image representation of the point cloud generated by Bewley et al.~\cite{bewley2020range}. From this 64$\times$2048 range image, we crop a central 64$\times$512 patch corresponding to the camera field of view, and treat this as an image into the DeepLab model. To generate per-scan groundtruth, we use nearest-neighbors to find the closest ground truth segmentation label to each point in the range image.

\paragraph{Cityscapes}
To test our method on a dataset without consistent instance labels and in a single-modal setting, we employ the Cityscapes dataset~\cite{Cordts_2016_CVPR}.
dataset. Cityscapes provides us with 2975, 500, and 1525 images of urban scenes for training, testing and validation, respectively. Furthermore, the dataset provides 8 ‘thing’ and 11 ‘stuff’ classes. In this work, we train our network on the `fine' set of training images with ground truth instance labels. In order to generate paired groundtruth, we use the unlabeled temporal frames, and randomly sample frames from the images 2 frames before or after the labeled image. We then warp the groundtruth from the labeled frame to each temporal frame using a UFlow~\cite{49244} model trained on the Waymo Open Dataset~\cite{sun2019scalability}, consisting of ~200,000 training images. To filter potential errors in the flow warping due to occlusions and other errors, we ignore warped labels based on the occlusion map generated by the flow model.
For Cityscapes, we rely on a pre-trained semantic segmentation network for assigning class IDs, in particular an implementation of Panoptic Deeplab~\cite{cheng2020panoptic}, with an Xception-71 backbone, trained on Cityscapes. Our implementation has a mIOU of 68.8\% on the Cityscapes validation set. As embeddings along the boundary are particularly challenging, and the Cityscapes labels are typically reliable, we dedicate half of the points when sampling in the contrastive loss to embeddings that are within $b=10$ pixels of the boundary of each instance mask.

\subsection{Results on KITTI-360}
\begin{figure}[t]
  \begin{center}
  \frame{
  \includegraphics[trim={-10pt 0 5pt 0},clip,width=1.0\linewidth]{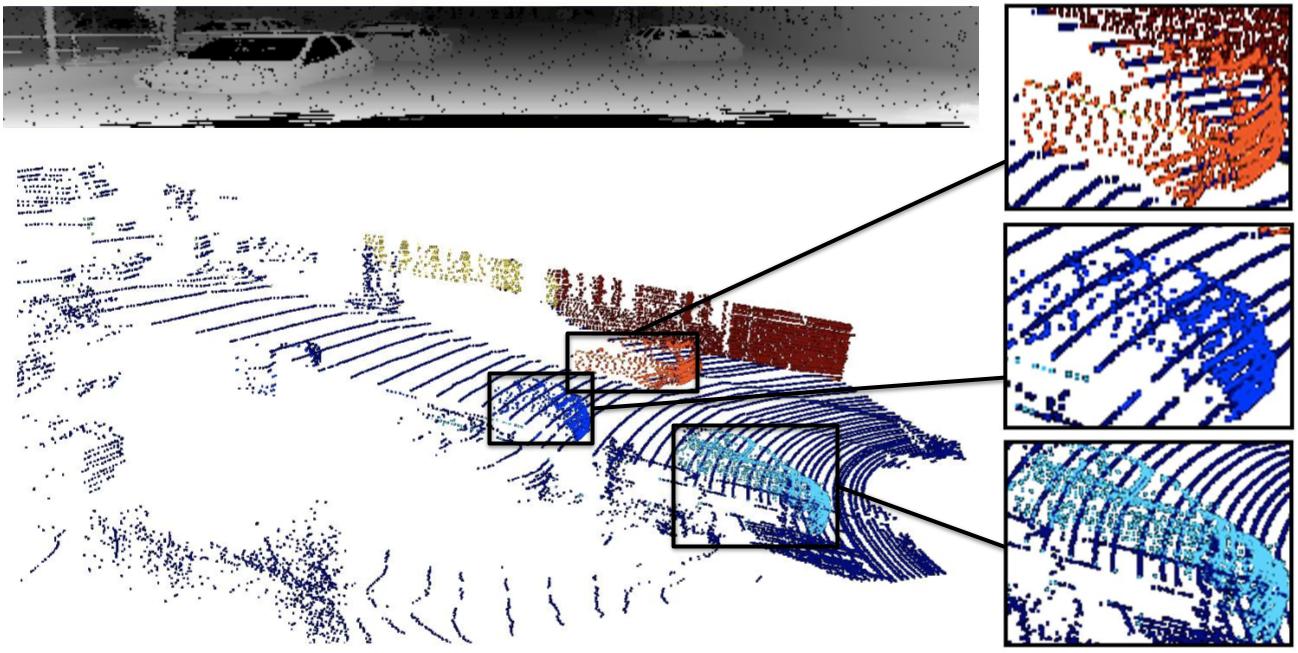} 
  }
  \end{center}
  \vspace{-3mm}  %
  \caption{Close-up analysis of our 3D LiDAR segmentation network, operating on range images (top-left), demonstrating accurate point-cloud instance segmentation.}
  \label{fig:kitti_pointcloud}   
\end{figure}

In Table~\ref{tab:kitti_ablation}, we report an ablation study on the effects of adding additional signals for the contrastive loss, both temporally and from different modalities. Overall, we observe that instance mAP for images improves as we add each additional frame to the consistency loss. The overall trend that we observe is that the network improves significantly for rare classes such as bicycle, motorcycle, rider and truck, while regressing slightly for the more common car class. This results in a significant overall mAP improvement. Note that, because the ablation models are trained on a much smaller training set, the mAP numbers in these experiments are naturally lower than those for the test set. We show qualitative examples from our best model (Temporal RGB + Temporal LiDAR) in Figure~\ref{fig:kitti_qualitative}, where we demonstrate that the networks are able to predict embeddings that are consistent for objects across the sensor modalities. We also show our LiDAR instance segmentation results projected to 3D in Figure~\ref{fig:kitti_pointcloud}, demonstrating that the network can generate accurate 3D instance segmentation when operating only on the range image. However, we did not observe similar improvements in 3D LiDAR mAP when training with the proposed cross-modality contrastive loss. Our observation is that, as the 2D labels are generated by projecting the 3D labels (and applying a CRF), they are typically much noisier than the 3D labels, and cannot provide useful signal for the 3D part of the model.

In Table~\ref{tab:kitti_test}, we report a comparison in the mAP of our final method against the popular MaskRCNN method~\cite{he2017mask} on the held-out test set, where we outperform the ResNet-50 baseline, while approaching the performance of the ResNet-101 baseline. At the time of writing, the validation set for this dataset was not available, and better results will be achievable once the same training splits are available as for the baselines.

\begin{table}[h]
\small
\centering
\begin{tabular}{l|c}
\specialrule{.1em}{.05em}{.05em} 
Method & AP ($\%$)\\
\specialrule{.1em}{.05em}{.05em}
Mask-RCNN ResNet-50 & 19.5 \\
Mask-RCNN ResNet-101 & 20.9 \\
Ours with temporal + lidar temporal & 20.3 \\
\hline
\end{tabular}
\caption{Results on the KITTI-360 held out test set.}
\label{tab:kitti_test}
\end{table}

\subsection{Results on Cityscapes}
 
In Table~\ref{tab:results_cityscapes}, we report our instance segmentation results on Cityscapes compared to existing methods. The reported results were obtained on the Cityscapes validation set. To provide a fair comparison of the main methods themselves, we use the reported numbers for each method without test time augmentation. In addition, we report mAP for our method using groundtruth semantic segmentations in Table~\ref{tab:results_cityscapes_groundtruth}, and compare against past work which has used a similar metric. This approach allows us to evaluate the instance segmentation network independently from the performance of the semantic segmentation. Overall, the improvement from our cross-modal consistency loss allows us to strongly outperform bottom up methods such as Brabandere et al.~\cite{Brabandere2017SemanticIS} and Neven et al.~\cite{neven2019instance} with groundtruth semantics. It also pushes our embedding-based method close to the previous bottom up works such as SSAP~\cite{journals/corr/abs-1909-01616}. However, this work requires a complex post-processing procedure which has a significant latency, as reported in Table~\ref{tab:results_cityscapes}.
We underperform compared to Panoptic DeepLab, but note that this method, while being single shot, requires the prediction of object centers. These centers serve as top-down proposals requiring instances to have distinguishable centers and some form of NMS.

\begin{figure}[h!]
  \begin{center}
  \includegraphics[width=\linewidth]{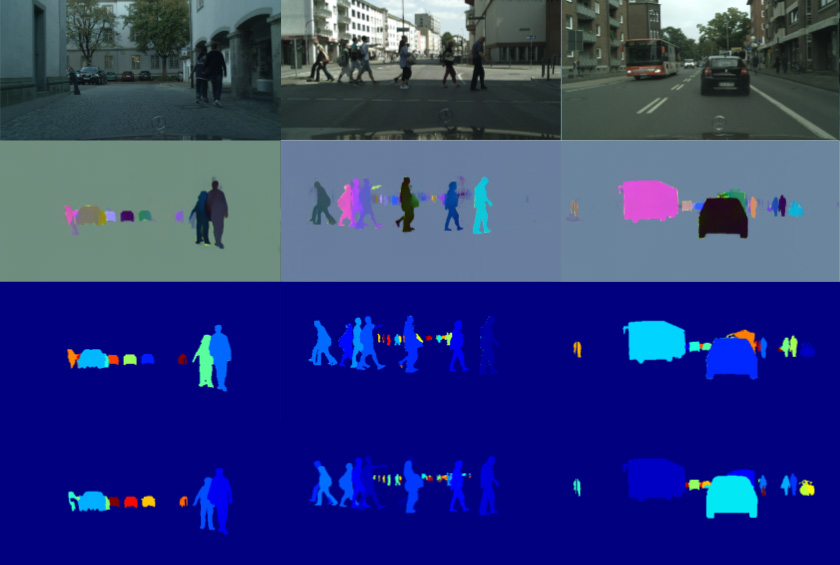} 
  \end{center}
  \vspace{-3mm}  %
  \caption{Visualization of per-pixel embedding vectors on Cityscapes: our method allows us to obtain per-pixel embedding vectors from RGB input images (first row). The embeddings robustly and precisely disentangle object instances across different classes (third row). To visualize the embeddings, we project them into RGB color space (second row). In the last row, we show the ground truth instance masks. Please note that the color of the predicted and ground truth masks are not supposed to match.
  }
  \label{fig:instance_embeddings}   
\end{figure}

We also show visualizations of our predictions in Figure~\ref{fig:instance_embeddings}, where the embeddings can robustly and precisely represent a large variety of object instances. 

\begin{table} [t]
\vspace{2mm}
\small
\centering
\scalebox{0.70}{
    \begin{tabular}{l|c|c|c|c}
    \specialrule{.1em}{.05em}{.05em} 
    Method & Approach & Input Size & AP (\%) & Speed (ms) \\
    \specialrule{.1em}{.05em}{.05em}

    Panoptic FPN~\cite{kirillov2019panopticfpn} & TD  & 512$\times$ 1024 & 33.2 & - \\
    UPSNet~\cite{Xiong_2019_CVPR}               & TD  & 1024$\times$2048 & 33.3 & 202ms \\
    Seamless~\cite{Porzi_2019_CVPR}             & TD  & 1024$\times$ 2048 & 33.6 & - \\
    Panoptic Deeplab~\cite{cheng2020panoptic} Xception-71  & TD$^1$ & 1025$\times$2049 & 35.3 & 175ms \\
    \hline
    SSAP~\cite{journals/corr/abs-1909-01616}    & BU  & 1024$\times$ 2048 & 31.5 & 260ms+$^2$ \\%
    
    Ours base                      & BU-C & 1024$\times$ 2048 & 29.0 & 182ms \\
    Ours temporal                              & BU-C &1024$\times$ 2048 & 30.9 & 182ms\\
    \hline
    \end{tabular}
    }
\caption{Cityscapes validation set results, without any test time augmentation or additional training data for all methods. TD: Top-Down. BU: Bottom-Up. BU-C: Bottom-Up via Contrastive Learning. $^1$Panoptic Deeplab is a single stage network, but relies on top-down object proposals of object centers. $^2$Reported time is for post-processing only and does not include the network forward pass.}
    \label{tab:results_cityscapes}
\end{table}  

\begin{table} [t!]
\vspace{1mm}
\small
\centering
\scalebox{0.71}{
    \begin{tabular}{l|c|c|c|c}
    \hline
    Method & Approach & Input Size & AP (\%) & Speed (ms) \\
    \specialrule{.1em}{.05em}{.05em}
    Brabandere et al.~\cite{Brabandere2017SemanticIS} (ResNet-38)     & BU-C & 384$\times$ 768 & 29.0 & 200ms \\
    Neven et al.~\cite{neven2019instance}                       & BU & 1024$\times$ 2048 & 40.5 & 91ms\\
    Ours base                               & BU-C & 1024$\times$ 2048 & 48.9 & 182ms \\
    Ours temporal                          & BU-C& 1024$\times$ 2048 & 50.6 & 182ms \\
    \hline
    \end{tabular}
    }
\caption{Cityscapes validation set results using groundtruth semantics for assigning classes to each instance. BU: Bottom-Up. BU-C: Bottom-Up via Contrastive Learning. For this setup, we significantly outperform other contrastive learning methods.}
\label{tab:results_cityscapes_groundtruth}
\end{table}

\textbf{Temporal Constraints:} To validate the impact of the temporal contrastive loss, we trained our model with and without contrasting over time (see Table~\ref{tab:results_cityscapes}, Ours base vs Ours temporal). From these results, introducing the contrastive loss over time has a significant impact on the predicted instances masks. Adding temporal frames to the contrastive loss improves overall AP by 2\%. The visualization of the predicted embeddings in Figure~\ref{fig:temporal_coherence} also shows fewer artifacts.

We also compute error metrics on the stability of the instance embeddings over time for the Cityscapes dataset. Given two frames, we warp the predicted embeddings of the next frame to the current frame, and compute the cosine distance between the average embedding of each instance in the two time steps, while ignoring any pixels estimated as `occlusion' from the optical flow. In addition, we compute as accuracy the proportion of instance pairs with cosine distance less than the clustering threshold $m=0.1$. These results can be found in Table~\ref{tab:results_temporal}. We observe that the majority of embeddings (75.1\%) are stable over time, even without the proposed temporal contrastive loss. However, there are a number of cases, such as large motions and dense objects, where the embeddings are not stable. Overall, our proposed loss reduces the cosine distance between instances by 0.025, and improves accuracy by 8.3$\%$ to 83.4$\%$.

\begin{table} [t!]
\vspace{1mm}
\small
\centering
\scalebox{1.0}{
    \begin{tabular}{l|c|c}
    \specialrule{.1em}{.05em}{.05em} 
    Method & Cosine Distance $\Downarrow$ & Accuracy $\Uparrow$ \\
    \specialrule{.1em}{.05em}{.05em}
    Without Temporal Loss & 0.087 & 75.1\\
    With Temporal Loss & 0.062 & 83.4 \\
    \hline
    \end{tabular}
    }
\caption{Temporal consistency metrics. We compute the cosine distance (in $[0, 1]$) between the average embedding for each instance between two temporal frames. Accuracy is the percentage of instances with cosine distance  $< m=0.1$ over time. Training with temporal consistency significantly increases performance.}
\label{tab:results_temporal}
\end{table}

\begin{figure}[t]
  \begin{center}
  \includegraphics[width=\linewidth]{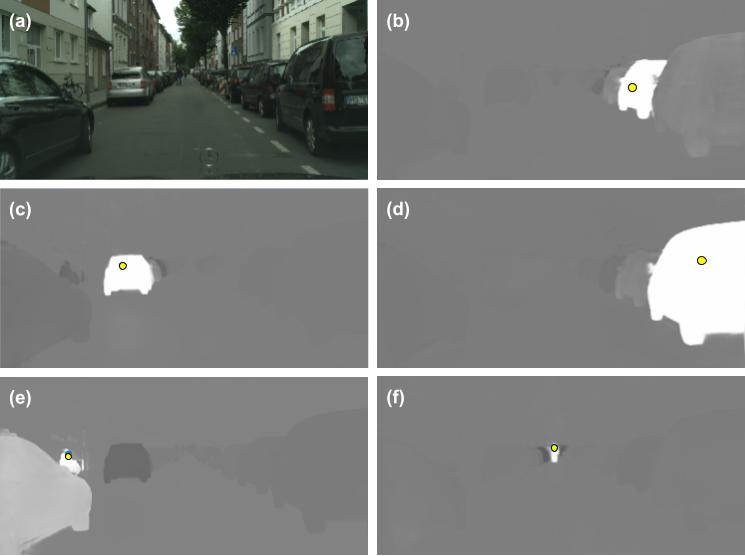} 
  \end{center}
  \vspace{-3mm}  %
  \caption{Pixel similarity: for an image with objects of different classes (a) we randomly select different pixel positions in the image (yellow dot) and show the similarity of the selected embedding to all other embeddings as gray-scale values. As shown, the learned embeddings allow us to meaningfully disentangle object instances, such as various different cars (b, c, d), intricate objects like bicycles~(e), and small-scale objects, such as pedestrians~(f).}  
  \label{fig:embedding_similarity}   
\end{figure}

\textbf{Embedding Similarity:} In Figure~\ref{fig:embedding_similarity} we show the results of an embedding similarity experiment. We select pixels in the image and compute the distance, shown in grayscale, of the selected embeddings with all other embeddings in the image. As illustrated, the learned embeddings allow us to disentangle object instances of the same class, across different classes, and the background. Our simple clustering scheme can generate high quality masks by thresholding this similarity.

\section{Limitations and Future Work} The main contribution of this method is the contrastive learning of instance embeddings between sensors and over time. As it stands, we believe that the learned embeddings are strong, but have chosen a relatively simple method to generate instance masks for evaluation. We use this post-processing to highlight the improvements in the embeddings themselves, where a more complex method may obscure such improvements. However, there is room for improvement if stronger AP scores are desired, by adding a graph cut optimization such as in SSAP~\cite{journals/corr/abs-1909-01616} or a transformer head on top of the embeddings as in MaX-DeepLab~\cite{wang2021max}.

\section{Conclusion}
We have introduced a novel method for learning pixel-wise embeddings as a representation for individual object instances. Compared to other bottom-up approaches, we employ a contrastive learning scheme; embeddings of the same object instance are pushed closer together, while they are pulled away from other objects and  the background. In particular, we leverage information between sensors, as well as over time, by applying our cross-modal contrastive loss on the union of all predicted embeddings for a given scene. Through quantitative experiments, we show that this contrastive loss is able to significantly improve instance segmentation performance, and coherent instance embeddings between sensors and time, which are clustered to generate high quality instance masks.

{\small
\bibliographystyle{ieee_fullname}
\bibliography{main}
}

\end{document}